\documentclass{article}

\usepackage{microtype}
\usepackage{graphicx}
\usepackage{booktabs} % for professional tables

\usepackage{hyperref}

\usepackage[accepted]{icml2020}

\usepackage{amsmath}
\usepackage{amssymb}
\usepackage{amsthm}
\usepackage{csquotes}
\usepackage{subcaption}
\usepackage{comment}

\icmltitlerunning{Some Insights into Lifelong Reinforcement Learning Systems}

\begin{document}

\twocolumn[
\icmltitle{Some Insights into Lifelong Reinforcement Learning Systems}




\begin{icmlauthorlist}
\icmlauthor{Changjian Li}{uw}
\end{icmlauthorlist}

\icmlaffiliation{uw}{Department of Electrical and Computer Engineering, University of Waterloo, Canada}

\icmlcorrespondingauthor{Changjian Li}{changjian.li@uwaterloo.ca}

\icmlkeywords{Machine Learning, ICML}

\vskip 0.3in
]



\printAffiliationsAndNotice{}  % leave blank if no need to mention equal contribution

\begin{abstract}
A lifelong reinforcement learning system is a learning system that has the ability to learn through trail-and-error interaction with the environment \emph{over its lifetime}. In this paper, I give some arguments to show that the traditional reinforcement learning paradigm fails to model this type of learning system. Some insights into lifelong reinforcement learning are provided, along with a simplistic prototype lifelong reinforcement learning system.
\end{abstract}

\section{Introduction}

An \emph{agent} is an abstraction of a decision-maker. At each time instance $t$, it receives an observation $o_{t} \in O$, and outputs an action $a_{t} \in A$ to be carried out in the environment it lives in. Here, $O$ is the (finite) set of possible observations the agent can receive, and $A$ is the (finite) set of actions the agent can choose from. An agent's observation $o_{t}$ depends on the current environment state $s_{t} \in S$ through an agent observation function $S \to O$, where $S$ is the set of possible environment states. The \emph{observation history} $h^{o}_{t} = (o_{1}, o_{2}..., o_{t})$ is the sequence of observations the agent has received \emph{till time $t$}. Let $H^{o}_{t}$ be the set of possible observation histories of \emph{length} $t$, the \emph{policy $\pi_{t}: H^{o}_{t} \to A$ at time $t$} is defined as the mapping from an observation history of length $t$ to the action the agent will take. An agent's behavior can thus be fully specified by its policy across all timesteps $\pi = (\pi_{1}, \pi_{2}, ..., \pi_{t}, ...)$. Throughout the paper, it is assumed that an agent has a finite lifespan $T$.

\subsection{Scalar Reward Reinforcement Learning System}
We are interested in agents that can achieve some goal. In reinforcement learning, a goal is expressed by a scalar signal $r_{t} \in \mathbb{R}$ called the \emph{reward}. The reward is dependent on the agent's observation history, and is assumed to be available to the agent at each timestep in addition to the observation $o_{t}$. Our aim is to find policies that maximize the expected cumulative reward an agent receives over its lifetime:
\begin{align}\label{eq:rl}
    \max_{\pi} \mathbb{E}[\sum_{t=1}^{T} {r_{t}(h^{o}_{t})}]
\end{align}
Using the maximization of expected cumulative scalar reward to formulate the general notion of goal is a design choice in reinforcement learning, based on what is commonly known as the \emph{reward hypothesis}~\cite{Sutton1998}, In Sutton's own words:
\begin{displayquote}
That all of what we mean by goals and purposes can be well thought of as the maximization of the expected value of the cumulative sum of a received scalar signal (called reward).
\end{displayquote}{}

This design choice, however, is somewhat arbitrary. Among other things, the reward needs not be a scalar (e.g. multi-objective reinforcement learning~\cite{WHITE1982639}), nor does it have to be a quantity whose cumulative sum is to be maximized (which we will come to shortly). Leaving aside the question of whether or not all goals \emph{can} be formulated by Eq. \ref{eq:rl}, I intend to show in this paper that the problem of lifelong reinforcement learning probably \emph{should not} be formulated as such.

Note that in Eq. \ref{eq:rl}, I defined the reward in terms of the observation history, instead of the history of environment states as in most reinforcement learning literature. This reflects the view that reward signals are internal to the agent, as pointed out by Singh et al.~\yrcite{DBLP:conf/nips/SinghBC04} in their work on intrinsic motivation. Since the observations are all that the agent has access to from the external environment, the intrinsic reward should depend on the environment state \emph{only} through the agent's observation history.

\begin{figure}
    \centering
    \includegraphics[width=0.4\textwidth]{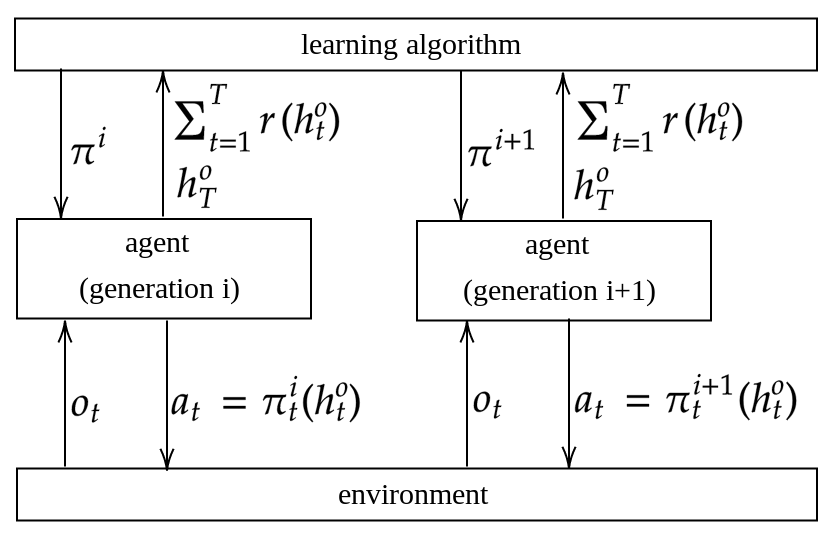}
    \caption{Architecture of a traditional reinforcement learning system. At the beginning of an agent's life, it receives a policy $\pi^{i}=(\pi^{i}_{1}, \pi^{i}_{2}, ... \pi^{i}_{T})$ from the learning algorithm that carries out a mix of exploitation and exploration, where the superscript $i$ indicates that the agent belongs to the $i$th generation. The agent receives an observation $o_{t}$ at each timestep $t$, and act according to $\pi^{i}_{t}$. At the end of the agent's life, the learning algorithm gathers the observation history $h^{o}_{T}$ and the cumulative reward $\sum_{t=1}^{T}{r(h^{o}_{t})}$ from the agent, and outputs the the next policy $\pi^{i+1}$ to be executed. The learning algorithm does not need to optimize the performance of any particular $\pi^{i}$, as long as it is guaranteed to be able to eventually find the policies that maximize the expected cumulative reward.}
    \label{fig:standardRL}
\end{figure}

Although the above reinforcement learning formulation recognizes the reward as a signal intrinsic to an agent, it focuses on learning across different generations~\footnote{Usage of the word `generation' here is only to emphasize that learning cannot be achieved within an agent's lifespan, and does not imply that evolution algorithms need to be used.} of agents, as opposed to learning within an agent's lifespan. From an agent's point of view, the cumulative reward is known only when it reaches its end of life, by which time no learning can be done by the `dying' agent itself. The individual reward received at each timestep does not really matter, since the optimization objective is the cumulative sum (of reward). The information gathered by the agent, however, can be used to improve the policy of the next generation. In other words, with the conventional reinforcement learning formulation, learning can only happen at a level higher than the lives of individual agents (Figure \ref{fig:standardRL}), with the goal that an optimal agent can eventually be found --- the lifetime behavior of a particular agent is not of concern. 

{}

\subsection{Towards Lifelong Reinforcement Learning} \label{sec:lifelong}
In lifelong reinforcement learning, on the other hand, the focus is the agent's ability to learn and adapt to the environment throughout its lifetime. Intuitively, this implies that learning component of the learning system should reside within the agent. 

To shed some lights on lifelong reinforcement learning, consider the Q-learning~\cite{DBLP:journals/ml/WatkinsD92} algorithm for the standard reinforcement learning problem formulated by Eq. \ref{eq:rl}. For the purpose of this example only, it is further assumed that:
\begin{itemize}
\item The reward depends only on the current observation. I.e., $r(h^{o}_{t}) = r(o_{t})$
\item Observations are Markov with respect to past observations and actions. I.e., $P(o_{t}|o_{t-1}, a_{t-1}, ..., o_{1}, a_{1}) = P(o_{t}|o_{t-1}, a_{t-1})$
\end{itemize}
These assumptions are only made so that Q-learning will find the solution to Eq. \ref{eq:rl}, and are not essential for the general discussion. The (non-lifelong) learning system works as follows:
\begin{enumerate}
\item The agent receives its initial Q estimate from the past generation.
\item At each timestep $t$, the agent takes an $\epsilon$-greedy action based on the current Q estimate, then does a Bellman update on the Q estimate:
\begin{align}\label{eq:bellman}
\begin{split}
    &Q(o_{t}, a_{t}) := \\ 
    &Q(o_{t}, a_{t}) + \alpha (r(o_{t}) + \max_{a} Q(o_{t+1}, a) - Q(o_{t}, a_{t}))
\end{split}
\end{align}{}
\item When the agent dies, pass the updated Q estimate to the next generation.
\end{enumerate}

At first sight, the fact that the Q estimate is updated every timestep seems to contradict my argument that learning only happens across generations. However, for Eq. \ref{eq:bellman} to be a valid update, the timestep $t$ needs to be part of the observation --- the observation $o_{t}$ here is in fact the raw observation $o^{-}_{t}$ augmented by time $t$, i.e., $o_{t}=(o^{-}_{t}, t)$. Since the timestep is part of the observation, no same observation will be experienced more than once throughout the agent's lifetime, and it makes no difference \emph{to the agent} whether the Q estimate is updated every timestep, or after its life ends~\footnote{The statement does not strictly hold true if function approximation is used. An update to $Q_\theta(o_{t}, a)$ can potentially affect the Q estimate of all other observations. However, this is more a side effect than a desired property.}.

It's clear that for an agent to exhibit any sensible behavior, the initial Q estimate it inherits from the past generation is vital. If the agent receives a random initial Q estimate, then it's lifelong behavior is bound to be random and meaningless. On the other side of the spectrum, if the agent receives the true Q function, then it will behave optimally. This suggests that if we care about the lifetime behaviour (which includes lifelong learning behavior) \emph{of a Q-learning agent}, then $Q(o_{t}, \cdot)$ is a fundamental signal the agent needs to receive in addition to the scalar reward. In a sense, if the signal represented by the scalar reward is a specification of what the goal \emph{is}, then the signal represented by the Q estimate is the knowledge past generations have collected about what the goal \emph{means} for this type of agent. As an analogy, the pain associated with falling to the ground could be the former signal, while the innate fear of height could be the latter. 

\begin{figure}
    \centering
    \includegraphics[width=0.4\textwidth]{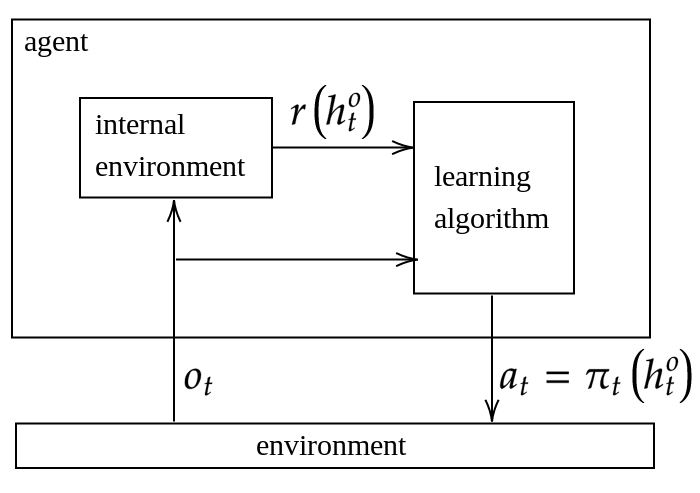}
    \caption{Architecture of lifelong reinforcement learning system. In contrast to traditional reinforcement learning (Figure \ref{fig:standardRL}), the learning algorithm resides inside the agent. The internal environment of the agent can be thought of as a built-in mechanism for the agent-designer to communicate with the agent (through the reward). At each timestep, the learning algorithm receives some message (encoded in the form of reward $r(h^{o}_{t})$) from the agent's internal environment, and outputs a policy $\pi_{t}$ as a response.}
    \label{fig:lifelong}
\end{figure}

From a computational perspective, the separation of these two signals may not be necessary. Both signals can be considered as `annotations' for the observation history that the agent receives along with its observation, and can be incorporated into the concept of reward. The reward signals are no longer restricted scalars, nor are they necessarily quantities whose cumulative sum is to be maximized --- they are just messages in some \emph{reward language} that `encode' the knowledge pertaining to an agent's observation history --- knowledge that enables the agent to learn continuously throughout its life. Such knowledge may include the goals of the agent, the subgoals that constitute these goals, the heuristics for achieving them, and so on. The reward is then `decoded' by the learning algorithm, which defines how the agent responds to the reward given the observation history. The learning system should be designed such that by responding to the reward in its intended way, the agent will learn to achieve the goals implied by the reward before its end of life (Figure \ref{fig:lifelong}).

To be precise, the reward $r(h^{o}_{t}) \in \Sigma$ now belongs to some \emph{reward space} $\Sigma$. The \emph{learning algorithm} is a mapping from reward histories to policies. Denoting the set of possible reward history of length $t$ as $H^{r}_{t}$, and the set of all possible policies at time $t$ as $\Pi_{t}$, the learning algorithm $m$ can be represented by $m=(m_{1}, m_{2}, ..., m_{t}, ..., m_{T})$, where $m_{t}: H^{r}_{t} \to \Pi_{t}$. The formulation is general, and a learning system formulated as such is not automatically a lifelong learning system. In fact, it subsumes traditional reinforcement learning: the reward space is set to the real numbers ($\Sigma = \mathbb{R}$), and the learning algorithm can be set to any algorithm that converges to a policy that maximizes the expected cumulative reward. Unfortunately, the reward in traditional reinforcement learning does not contain enough information for an agent to learn \emph{within} its lifetime.

{}

Viewing the reward as a general language, and the learning algorithm as the response to the reward opens up the possibilities for principled ways to embed learning bias such as guidance and intrinsic motivation into the learning system, instead of relying solely on manipulating the scalar reward on an ad-hoc basis. In the rest of the paper, my focus remains on lifelong reinforcement learning, more specifically, what lifelong reinforcement learning requires of the reward language and the corresponding learning algorithm. 

\subsection{Reward as Formal Language}
\label{sec:automaton}
Although the term `language' used above can be understood in its colloquial sense, it can also be understood as the formal term in automata theory. To see this, consider the following \emph{deterministic finite automaton} $\langle \Sigma, Q, \delta, q_{0}, F \rangle$, where:
\begin{itemize}
    \item $\Sigma$ is the \emph{alphabet} of the automaton, and is set to the reward space of the learning system. In other words, the alphabet of this automaton consists of all possible reward the agent can receive at any single timestep. A \emph{string} is a sequence of symbols chosen from some alphabet. For this particular automaton, a string is in fact a sequence of reward, so the notation for reward history $h^{r}_{t}$ is also used to denote a string of length $t$. The set of all strings of length $k$ over $\Sigma$ is denoted as $\Sigma^{k}$, and the set of all strings (of any length) is denoted as $\Sigma^{*}$.
    \item $Q$ is the set of states of the automaton. Each state of this automaton is a possible pair of reward history and policies till some timestep $t$. For example, members of $Q$ include: 
    \[\langle h^{r}_{t=1}, \quad (\pi_{1})\rangle\]
    \[\langle h^{r}_{t=2}, \quad (\pi_{1}, \pi_{2})\rangle\]
    \[...\]
    \[\langle h^{r}_{t=T}, \quad (\pi_{1}, \pi_{2}, ..., \pi_{T})\rangle\] 
    for any $\pi_{1} \in \Pi_{1}$, $\pi_{2} \in \Pi_{2}$, ..., $\pi_{T} \in \Pi_{T}$, and $h^{r}_{t=1} \in \Sigma^{1}$,  $h^{r}_{t=2} \in \Sigma^{2}$, ..., $h^{r}_{t=T} \in \Sigma^{T}$. In addition, Q has a special `empty' member $q_{0}$, which corresponds to the initial state before any reward is received.
    \item $\delta: (Q \times \Sigma) \to Q$ is the \emph{transition function}. The transition function corresponds to the learning algorithm of the learning system, so we have $\delta(\langle h^{r}_{t}, (\pi_{1}, ..., \pi_{t})\rangle, \quad r_{t+1}) = \langle h^{r}_{t+1}, \quad (\pi_{1}, ..., \pi_{t}, m_{t+1}(h^{r}_{t+1})) \rangle$, where $h^{r}_{t+1} = (h^{r}_{t}, r_{t+1})$.    
    \item $q_{0}$ is the initial state of the automaton as explained above.
    \item $F \subset Q$ is the set of \emph{accepting} states, which are the desired states of the automaton.
\end{itemize}{}
It's not hard to see that this automaton is a model of the learning system described in Section \ref{sec:lifelong}, with its desired property specified by the accepting states $F$. In this paper, the desired property is that the system be a lifelong learning system, so the accepting states $F$ are the set of $\langle h^{r}_{T}, \quad (\pi_{1}, \pi_{2}, ..., \pi_{T}) \rangle$ pairs that correspond to a lifelong learner~\footnote{Recall that an agent's behavior is fully decided by its policy $\pi = (\pi_{1}, \pi_{2}, ..., \pi_{T})$. Therefore given a reward history $h^{r}_{T}$, the policy is sufficient for us to tell whether the agent is a successful lifelong learner.}.

To specify learning objectives, each possible reward $r \in \Sigma$ is assigned some semantics. These semantics implicitly define the set of valid reward sequences $L \subset \Sigma^{*}$. Since $L$ is a subset of $\Sigma^{*}$, it is a language over $\Sigma$. We want to make sure that --- for all reward sequences in $L$, lifelong learning can be achieved by the learning system abstracted by this automaton, or equivalently, all reward sequences in $L$ lead to accepting states $F$.

\section{A Prototype Lifelong Reinforcement Learning System}
\label{sec:prototype}
Designing a lifelong reinforcement learning system involves designing the reward language and the learning algorithm holistically. Intuitively, the reward needs to contain enough information to control the relevant aspects of the learning algorithm, and the learning algorithm in turn needs to `interpret' the reward signal in its intended way. In this section, I aim to provide some insights into the design process with a prototype lifelong reinforcement learning system.

\subsection{Reward Language}
The main reason lifelong learning is impossible in conventional reinforcement learning is that the learning objective in conventional reinforcement learning is \emph{global}, in the sense that the goal of the agent is defined in terms of the observation history of its entire life. For a lifelong reinforcement learning agent, the learning objectives should instead be \emph{local}, meaning that the goals should be defined only for some smaller tasks that the agent can encounter multiple times during its lifetime. Once a local goal \emph{expires}, whether it is because the goal has been achieved or because the agent has failed to achieve it within a certain time limit, a new local goal (can potentially be another instantiation of the same goal) ensues. This way, the agent has the opportunity to gather knowledge for each of the goals, and improve upon them, all within one life. Local goals like this are ubiquitous for humans. For example, when a person is hungry, his main concern is probably not the global goal of being happy for the rest of his life --- his goal is to have food. After the person is full, he might feel like taking a nap, which is another local goal. In fact, the local goals and the transition of them seems to embody what we mean by intrinsic motivation.

To be able to specify a series of local goals, the reward in this prototype learning system has two parts: the \emph{reward state} $r^{s}_{t} \in G$, and the \emph{reward value} $r^{v}_{t} \in \mathbb{R}$, where $G$ is the set of local goals the agent may have. This form of reward is inspired by the \emph{reward machine}~\cite{DBLP:conf/icml/IcarteKVM18}, a Mealy machine for specifying history-dependent reward, but the semantics we assign to the reward will be different. Also note that this Mealy machine bears no relation to the automaton we discussed in Section \ref{sec:automaton} --- the reward machine models the reward, while the automaton in Section \ref{sec:automaton} models the learning system, and takes the reward as input. Each reward state $r^{s}$ corresponds to a local goal. When a local goal (or equivalently, a reward state) expires, the agent receives a numerical reward value $r^{v}$. For all other timesteps (other than the expiration of local goals), the reward value can be considered to take a special \texttt{NULL} value, meaning that no reward value is received. The reward value is an evaluation of the agent's performance in an episode of a reward state, where an \emph{episode} of a reward state is defined as the time period between the expiration of the previous reward state (exclusive) and the expiration of the reward state itself (inclusive). The reward state can potentially depend on the entire observation history, while the reward value can only depend on the observation history of the episode it is assessing. Overall, the reward is specified by $(r^{s}_{t}, r^{v}_{t}) = r(h^{o}_{t})$.

The local goals described here are technically similar to subgoals in \emph{hierarchical reinforcement learning}~\cite{DBLP:journals/jair/Dietterich00, DBLP:journals/ai/SuttonPS99, DBLP:conf/nips/ParrR97}. However, the term `subgoal' suggests that there is some higher-level goal that the agent needs to achieve, and that the higher-level goal is the true objective the agent needs to optimize. That is not the case here --- although it is totally possible that the local goals are designed in such a way that some global goal can be achieved, the agent only needs to optimize the local goals. 

The reward language in this prototype system makes two assumptions on the learning algorithm. As long as the two assumptions are met, the learning algorithm is considered to `interpret' the reward correctly. The first assumption is that the learning algorithm only generates policies that are \emph{episode-wise stationary}, meaning that $\pi_{t_{1}} = \pi_{t_{2}}$ for any timesteps $t_{1}$ and $t_{2}$ in the same episode of a reward state, and that $\pi_{t_{1}}: O \to A$. This assumption is not particularly restrictive, because in cases where a local goal requires a more complex policy, we can always split the goal into multiple goals (by modifying the reward function) for which the policies are episode-wise stationary. With this assumption, we can use a single policy $\pi_{r^{s}}: O \to A$ to represent the policies at all timesteps within an episode of reward state $r^{s}$. The second assumption is that the learning algorithm keeps a pool of `elite' policies for each reward state: a policy that led to high reward value in some episode has the opportunity to enter the pool, and a policy that consistently leads to higher reward value eventually dominates the policy pool. The exact criterion for selection into the pool (e.g., to use the expected reward value as the criterion, or to use the probability of the reward value being higher than a certain threshold, etc.) is not enforced, and is left up to the learning algorithm.

\subsection{Learning Algorithm}
The learning algorithm in this prototype lifelong learning system is an \emph{evolutionary algorithm}, adjusted to meet the assumptions made by the reward. The algorithm maintains a policy pool $D_{r^{s}}$ of maximum size $d$ for each reward state $r^{s} \in G$. Each item in the pool is a two tuple $\langle \pi, r^{v}_{\pi} \rangle$ where $\pi$ is a policy and $r^{v}_{\pi}$ is the reward value of the last episode in which $\pi$ was executed. Conceptually, the algorithm consists of three steps: policy generation, policy execution, and (policy) pool update, which are described below.

\subsubsection*{Policy Generation}
When an episode of reward state $r^{s}$ starts, a policy $\pi_{r^{s}}$ is generated from one of the following methods with probability $p_{1}$, $p_{2}$, $p_{3}$, respectively: 
    \begin{enumerate}
        \item Randomly sample a policy from the policy pool $D_{r^{s}}$, and mutate the policy.
        \item Randomly sample a policy from $D_{r^{s}}$ and keep it as is. Remove the sampled policy from $D_{r^{s}}$. This is to re-evaluate a policy in the pool. Since the transition of observations might be stochastic, the same policy does not necessarily always result in the same reward value.
        \item Randomly generate a new policy $\pi_{r^{s}}: O \to A$ from scratch. This is to keep the diversity of the policy pool.
    \end{enumerate}{}
$p_{1}$, $p_{2}$ and $p_{3}$ should sum up to $1$, and are hyper-parameters of the algorithm.

\subsubsection*{Policy Execution}
Execute the generated policy $\pi_{r^{s}}$ until a numerical reward value $r^{v}$ is received.

\subsubsection*{Pool Update}
If the policy pool is not full, insert $\langle\pi_{r^{s}}, r^{v}\rangle$ into the pool. Otherwise compare $r^{v}$ with the minimum reward value in the pool. If $r^{v}$ is greater than or equal to the minimum reward value, replace the policy and reward value pair (that has the minimum reward value) with $\langle\pi_{r^{s}}, r^{v}\rangle$.

\subsection{Embedding Learning Bias}
\emph{Learning bias} in reinforcement learning systems refers to the explicit or implicit assumptions made by the learning algorithm about the policy. Our assumption that the policy is episode-wise stationary is an example of learning bias. Arguably, a good learning bias is as important as a good learning algorithm, therefore it is important that mechanisms are provided to embed learning bias into the learning system.

A straight-forward way to embed learning bias into the above lifelong learning system is through the policy generation process. This includes how existing policies are mutated, and what distribution new policies are sampled from. The learning bias provided this way does not depend on the agent's observation and reward history, and is sometimes implicit (e.g., the learning bias introduced by using a neural network of particular architecture).

\begin{figure}
    \centering
    \includegraphics[width=0.2\textwidth]{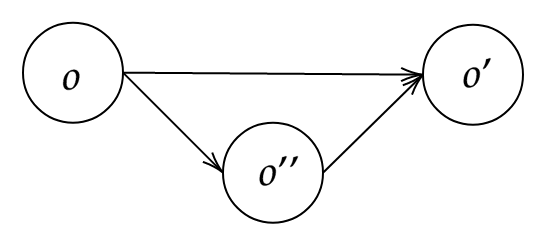}
    \caption{A simplistic abstraction of guidance in reinforcement learning.}
    \label{fig:guidance}
\end{figure}

Another type of learning bias common in reinforcement learning is \emph{guidance}, the essence of which can be illustrated by Figure \ref{fig:guidance}. Suppose in some reward state, the agent starts from observation $o$ and the goal is to reach~\footnote{For sake of terminological convenience, we pretend that the observations here are environment states.} observation $o'$. Prior knowledge indicates that to reach $o'$, visiting $o''$ is a good heuristic, but reaching $o''$ itself has little or no merit. In other words, we would like to encourage the agent to visit and explore around $o''$ more frequently (than other parts of the observation space) until a reliable policy to reach $o'$ is found.
{}
To provide guidance to the agent in the prototype lifelong learning system, we can utilize the property of the learning algorithm that policies leading to high reward values will enter the policy pool. Once a policy enters the pool, it has the opportunity to be sampled (possibly with mutation) and executed. Therefore, we just need to assign a higher reward value for reaching $o''$ (before the expiration of the reward state) than reaching neither $o'$ nor $o''$. Also important is the ability to control the extent to which region around $o''$ is explored. To achieve this, recall that the learning algorithm occasionally re-evaluates policies in the policy pool. If we assign a lower reward value for reaching $o''$ with some probability, we can prevent the policy pool from being overwhelmed only by policies that lead to $o''$. In other words, the reward value for reaching $o''$ should have multiple candidates. Let $r^{v}(\{O-\{o', o''\}\})$ denote the reward value for an episode where the agent reaches neither $o'$ nor $o''$, $r^{v}(o')$ denote the reward value for reaching $o'$, we can set the reward value $r^{v}(o'')$ for reaching $o''$ as:
\[
r^{v}(o'') := 
\begin{cases}
a, & \text{with probability $p$} \\
b, & \text{with probability $1-p$}
\end{cases}
\]
where $b < r^{v}(\{O-\{o', o''\}\}) < a < r^{v}(o')$. The probability $p$ controls the frequency region around $o''$ is to be explored compared the other parts of the observation space~\footnote{Note that the word `probability' here should be interpreted as the `long-run proportion', and therefore the reward value needs not be truly stochastic. E.g., we can imagine that the reward has a third component which is the state of a pseudo-random generator.}.

\section{Experiment}
Now we evaluate the behaviour of the prototype lifelong reinforcement learning system. The source code of the experiments can be found at \url{https://gitlab.com/lifelong-rl/lifelongRL_gridworld}

\subsection{Environment}
\begin{figure}
    \centering
    \includegraphics[width=0.25\textwidth]{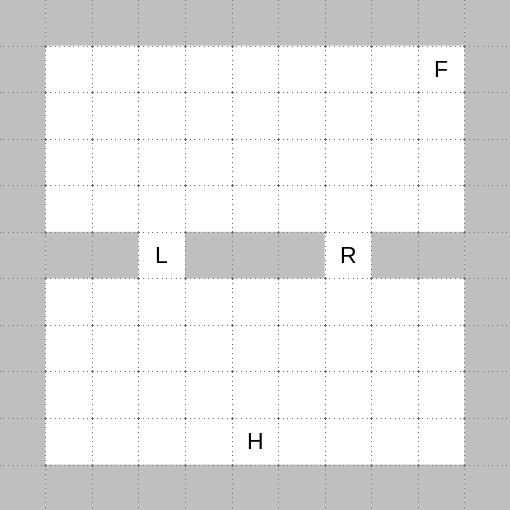}
    \caption{Gridworld environment.}
    \label{fig:gridworld}
\end{figure}

Consider a gridworld agent whose life revolves around getting food and taking the food back home for consumption. The agent lives in a $11$ by $11$ gridworld shown in Figure \ref{fig:gridworld}. The shaded areas are barriers that the agent cannot go through. Some potential positions of interest are marked with letters: \texttt{F} is the food source and is assumed to have infinite supply of food; \texttt{H} is the agent's home. To get to the food source from home, and to carry the food home, the agent must pass through one of the two tunnels --- the tunnel on the left is marked with \texttt{L} and the tunnel on the right is marked with \texttt{R}. At each timestep, the agent observes its position in the gridworld as well as a signal indicating whether it is in one of the four positions of interest (if yes, which), and chooses from one of the four actions: \texttt{UP}, \texttt{RIGHT}, \texttt{DOWN} and \texttt{LEFT}. Each action deterministically takes the agent to the adjacent grid in the corresponding direction, unless the destination is a barrier, in which case the agent remains in its original position. The agent starts from home at the beginning of its life, and needs to go to the food source to get food. Once it reaches the food source, it needs to carry the food back home. This process repeats until the agent dies. The lifespan of the agent is assumed to be $100$ million timesteps. The agent is supposed to learn to reliably achieve these two local goals within its lifetime.

{}

\subsection{Learning System Setup}
The reward state in this experiment is represented by the conjunction of Boolean variables. For example, if three Boolean variables $A$, $B$ and $C$ are defined, then the reward state would be in the form of $r^{s} = A \land B \land C$ or $r^{s} = A \land \neg B \land C$, etc. At the bare minimum, one Boolean variable $\mathtt{GET\_FOOD}$ needs to be defined for this agent, where $\mathtt{GET\_FOOD}$ being true corresponds to the local goal of going to the food source, and $\neg \mathtt{GET\_FOOD}$ corresponds to the local goal of carrying the food home. The agent receives a reward value of $+1$ if $\mathtt{GET\_FOOD}$ is true and the agent reaches $\mathtt{F}$, in which case the Boolean variable $\mathtt{GET\_FOOD}$ transitions to false. Similarly, the agent receives a reward value of $+1$ if $\neg \mathtt{GET\_FOOD}$ is true and the agent reaches $\mathtt{H}$, in which case $\mathtt{GET\_FOOD}$ transitions to true. On top of $\mathtt{GET\_FOOD}$, we define another Boolean variable $\mathtt{TIMED\_OUT}$, which indicates whether the agent has exceeded a certain time limit for trying to get to the food source, or for trying to carry the food home. If the reward state is $\neg \mathtt{TIMED\_OUT} \land \mathtt{GET\_FOOD}$, and the agent fails to reach $\mathtt{F}$ within the time limit, itreceives a reward value of $-1$, and the reward state transition to $\mathtt{TIMED\_OUT} \land \mathtt{GET\_FOOD}$. From $\mathtt{TIMED\_OUT}  \land \mathtt{GET\_FOOD}$, if the agent still fails to get to $\mathtt{F}$ within the time limit, it receives a reward value of $0$. The agent will remain in $\mathtt{TIMED\_OUT} \land \mathtt{GET\_FOOD}$, until it reaches $\mathtt{F}$, when the reward state transitions to $\neg \mathtt{TIMED\_OUT} \land \neg \mathtt{GET\_FOOD}$ (and receive a $+1$ reward value as already mentioned). For the case when $\mathtt{GET\_FOOD}$ is false, the reward transition is defined similarly. Throughout the experiments, the time limit is set to $24$, which is enough for the agent to accomplish any of the local goals. We refer to this reward design as the \emph{base} case.

Unfortunately, even for a toy problem like this, learning can be difficult if no proper learning bias is provided. Since there are $4$ actions and $74$ possible positions, the number of possible episode-wise stationary policies is $4^{74}$ for each reward state. Among those policies, very few can achieve the local goals. If the policy generation and mutation is purely random, it will take a long time for the agent to find a good policy.

\subsubsection*{Biased Policy}
The first learning bias we consider is \emph{biased policy}, which is in contrast to the \emph{unbiased policy} case where the policy generation and mutation is purely random. More specifically, we make the policy generation process biased towards policies that take the same action for similar observations. This would encourage policies that head consistently in one direction, and discourage those that indefinitely roam around between adjacent positions. 

\subsubsection*{Progress-Based Guidance}
The second learning bias we consider is guidance based on the agent's progress. Different from the base case where the agent always receives a $0$ (if $\mathtt{TIMED\_OUT}$ is true) or $-1$ (if $\mathtt{TIMED\_OUT}$ is false) reward value when it fails to achieve the local goal within the time limit, the agent now has some probability $p = 0.8$ of receiving a reward value proportional to the Manhattan distance $d$ it has traveled since the beginning of the episode. To be precise:
\[
r^{v} := 
\begin{cases}
0.01d & \text{with probability $p$} \\
\text{same as the base case} & \text{with probability $1-p$}
\end{cases}
\]
This way, policies leading to more progress (albeit not necessary towards the local goal) will be encouraged.

\subsubsection*{Sub-Optimal Guidance}
\label{sec:sub-optimal guidance}
Finally, we consider a case of sub-optimal guidance that encourages the agent to explore a sub-optimal trajectory. As we have mentioned, both reaching the food source from home and carrying the food home require the agent to go through one of the two tunnels. However, if the agent goes through the left tunnel, it has to travel more distance. Suppose that we prefer the agent to take the shorter route, but we only know the route that goes through the left tunnel; and as a result, we sub-optimally encourage the agent to explore the left tunnel. To guide the agent to take the left tunnel,  Boolean variable $\mathtt{VISITED\_LEFT}$ is introduced as an indicator of whether \texttt{L} has been visited since the last visitation of \texttt{F} or \texttt{H}. Now we have $2^{3} = 9$ elements in the reward space, corresponding to $9$ possible local goals. The reward transition is different from the base case in that if the agent has already visited \texttt{L} when the local goal $\mathtt{GET\_FOOD} \land \neg \mathtt{TIMED\_OUT} \land \neg \mathtt{VISITED\_LEFT}$ or $\neg \mathtt{GET\_FOOD} \land \neg \mathtt{TIMED\_OUT} \land \neg \mathtt{VISITED\_LEFT}$ times out, $\mathtt{VISITED\_LEFT}$ becomes true, and the agent will receive a reward value of $+0.6$ with $0.8$ probability, and $-0.2$ with $0.2$ probability. To express our preference for the shorter route, the agent receives a reward value of $+0.8$ (instead of $+1$) when it reaches \texttt{F} (when $\mathtt{GET\_FOOD}$ is true) or \texttt{H} (when $\mathtt{GET\_FOOD}$ is false) through the left tunnel. 

\subsection{Results}
\begin{figure}[h]
    \centering
    \begin{subfigure}[b]{0.5\textwidth}
        \centering
        \includegraphics[width=0.9\textwidth]{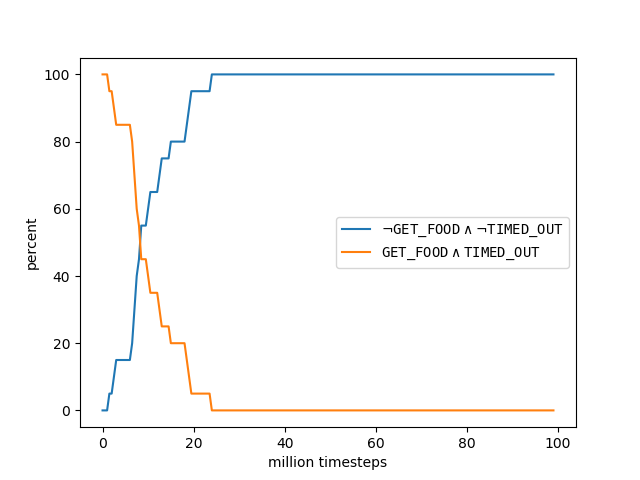}
        \caption{unbiased policy, progressed-based guidance}
    \end{subfigure}
    ~
    \begin{subfigure}[b]{0.5\textwidth}
        \centering
        \includegraphics[width=0.9\textwidth]{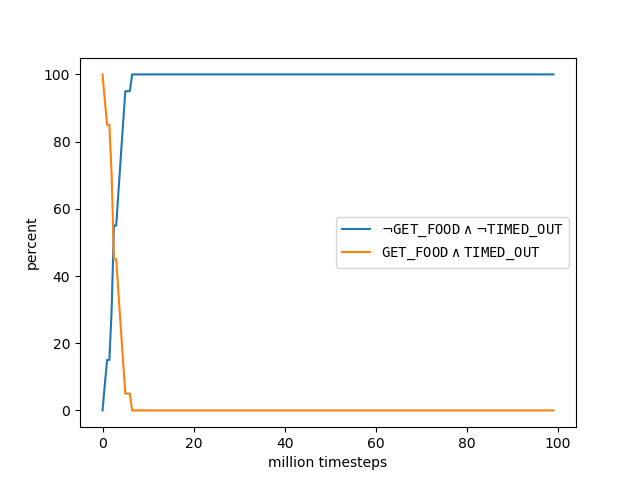}
        \caption{biased policy, progressed-based guidance}
    \end{subfigure}
    \caption{Learning curve for unbiased/biased policy with progress-based guidance, averaged over $20$ runs.}
    \label{fig:progress-guidance}
\end{figure}

Figure \ref{fig:progress-guidance} shows the learning curves for reward state $\mathtt{GET\_FOOD} \land \neg \mathtt{TIMED\_OUT}$ with progress-based guidance. The $x$-axis is the timesteps (in million), and the $y$-axis is the percentage of times the agent transitions into a particular next reward state starting from $\mathtt{GET\_FOOD} \land \neg \mathtt{TIMED\_OUT}$. A next reward state of $\neg \mathtt{GET\_FOOD} \land \neg \mathtt{TIMED\_OUT}$ means that the agent successfully reached \texttt{F} within the time limit, and a next reward state of $\mathtt{GET\_FOOD} \land \mathtt{TIMED\_OUT}$ means that the agent failed to do so. As we can see, with unbiased policy, it took the agent around $25$ million timesteps to achieve $100\%$ success rate; while with biased policy, this only took around $8$ million timesteps.

\begin{figure}[h]
    \centering
    \begin{subfigure}[b]{0.5\textwidth}
        \centering
        \includegraphics[width=0.8\textwidth]{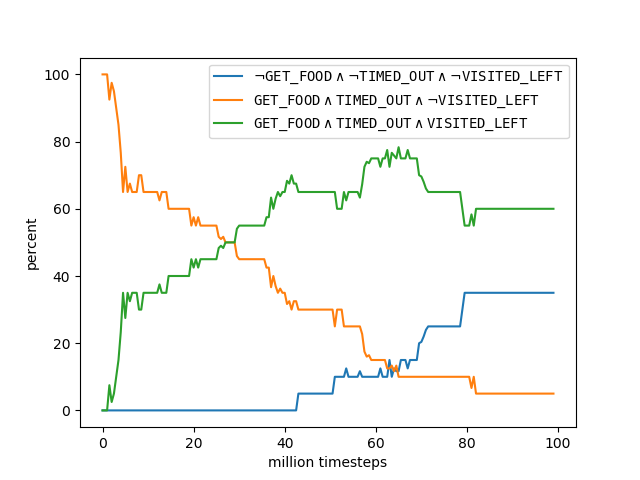}
        \caption{unbiased policy, sub-optimal guidance}
    \end{subfigure}
    ~
    \begin{subfigure}[b]{0.5\textwidth}
        \centering
        \includegraphics[width=0.8\textwidth]{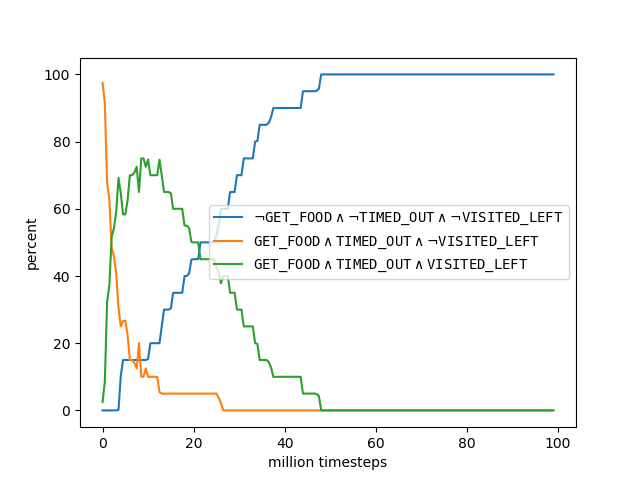}
        \caption{biased policy, sub-optimal guidance}
    \end{subfigure}
    \caption{Learning curve for unbiased/biased policy with sub-optimal guidance, averaged over $20$ runs.}
    \label{fig:sub-optimal_guidance}
\end{figure}

\begin{figure}[h]
    \centering
    \includegraphics[width=0.4\textwidth]{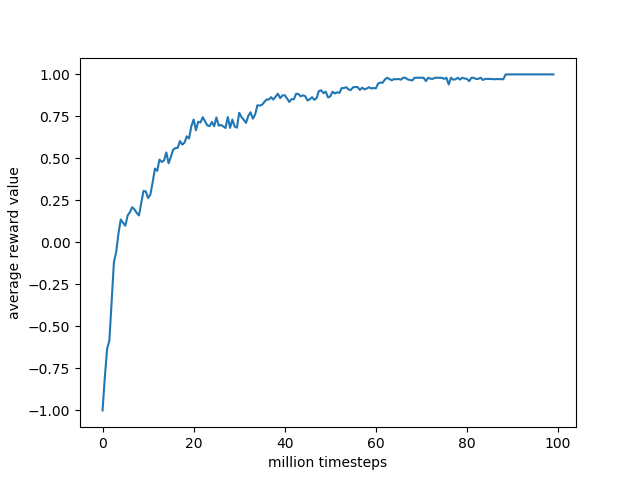}
    \caption{Reward value for $\mathtt{GET\_FOOD} \land \neg \mathtt{TIMED\_OUT} \land \neg \mathtt{VISITED\_LEFT}$ (biased policy with sub-optimal guidance, averaged over $20$ runs).}
    \label{fig:reward_value}
\end{figure}

Figure \ref{fig:sub-optimal_guidance} shows the learning curves for reward state $\mathtt{GET\_FOOD} \land \neg \mathtt{TIMED\_OUT} \land \neg \mathtt{VISITED\_LEFT}$ with the sub-optimal guidance described in Section \ref{sec:sub-optimal guidance}. Similar to Figure \ref{fig:progress-guidance}, the $x$-axis is the timesteps (in million), and the $y$-axis is the percentage of times the agent transitioned into a particular next reward state starting from $\mathtt{GET\_FOOD} \land \neg \mathtt{TIMED\_OUT} \land \neg \mathtt{VISITED\_LEFT}$. A next reward state of $\neg \mathtt{GET\_FOOD} \land \neg \mathtt{TIMED\_OUT} \land \neg \mathtt{VISITED\_LEFT}$ means that the agent successfully reached \texttt{F} within the time limit; a next reward state of $\mathtt{GET\_FOOD} \land \mathtt{TIMED\_OUT} \land \mathtt{VISITED\_LEFT}$ means that the agent failed to reach the food source, but was able to find a way to the left tunnel; and a next reward state of $\mathtt{GET\_FOOD} \land \mathtt{TIMED\_OUT} \land \neg \mathtt{VISITED\_LEFT}$ means that the agent was neither able to reach the left tunnel nor the food source within the time limit. As we can see, for both unbiased and biased policy, learning is much slower than progress-based guidance. This is likely due to the much sparser guidance signal --- the agent receives guidance only when it reaches the left tunnel. For the unbiased policy case, $100\%$ success rate was not achieved within $100$ million timesteps, but we can clearly see that exploration around the left tunnel was encouraged as intended. For the biased policy case, the agent was able to reach $100\%$ success rate after $50$ million timesteps. But was the agent able to figure out the optimal route, or did it only learn to take the sub-optimal route as guided? Recall that the agent receives a reward value of $+1$ if it takes the optimal route, and a reward value of $+0.8$ if it takes the sub-optimal route. As shown in Figure \ref{fig:reward_value}, although the agent was taking the sub-optimal route by $50$ million timesteps when it just learned to reach the food source reliably, it was eventually able to figure out the optimal route by $90$ million timesteps.

\section{Conclusions}
Lifelong reinforcement learning is sometimes viewed as a multi-task reinforcement learning problem~\cite{DBLP:conf/icml/AbelALL18}, where the agent must learn to solve tasks sampled from some distribution $\mathcal{D}$. The agent is expected to (explicitly or implicitly) discover the relation between tasks, and generalize its policy to unseen tasks from $\mathcal{D}$. The focus is therefore on the transfer learning~\cite{DBLP:journals/jmlr/TaylorS09} and continual learning~\cite{DBLP:books/sp/98/Ring98} aspects of lifelong reinforcement learning.

\begin{comment}
However, finding the connections between tasks is difficult: the same amount of reward might mean completely different things, and the same observation can have completely different implications. As a result, strong assumptions on the task distribution of tasks need to be made (e.g., maze tasks with different goal locations but the reward for reaching the goal is the same). 
\end{comment}

In this paper, I provided a systems view on lifelong reinforcement learning. In particular, I showed that the reward in a lifelong reinforcement learning system can be a general language, and that the language needs to be designed holistically with the learning algorithm. A prototype lifelong reinforcement learning system was given, with an emphasize on how learning bias can be embedded into the learning system through the synergy of the reward language and the learning algorithm.

%\begin{comment}
\section*{Acknowledgements}
The author would like to thank Gaurav Sharma (Borealis AI) for his comments on a draft of the paper.
%\end{comment}


%\clearpage
\bibliography{example_paper}
\bibliographystyle{icml2020}

\end{document}